\documentclass{article}

\usepackage[numbers,sort&compress]{natbib}
\usepackage[preprint]{neurips_2026}

\usepackage[utf8]{inputenc}
\usepackage[T1]{fontenc}
\usepackage{microtype}
\usepackage{graphicx}
\usepackage{subcaption}
\usepackage{booktabs}
\usepackage[hidelinks]{hyperref}
\usepackage{url}
\usepackage{amsmath}
\usepackage{amssymb}
\usepackage{amsfonts}
\usepackage{mathtools}
\usepackage{amsthm}
\usepackage{algorithm}
\usepackage{algorithmic}
\usepackage{nicefrac}
\usepackage{xcolor}
\usepackage{wrapfig}
\usepackage{multirow}
\usepackage{placeins}

\usepackage[capitalize,noabbrev]{cleveref}

\theoremstyle{plain}

\theoremstyle{definition}

\theoremstyle{remark}

\title{Whole-Brain Connectomic Graph Model Enables Whole-Body Locomotion Control in Fruit Fly}

\author{%
  Zehao Jin \\
  Tsinghua University \\
  \And
  Yaoye Zhu \\
  Tsinghua University \\
  \And
  Chen Zhang \\
  Tsinghua University \\
  \And
  Yanan Sui\thanks{Corresponding author: \texttt{ysui@tsinghua.edu.cn}} \\
  Tsinghua University \\
}

\begin{document}

\maketitle

\begin{abstract}
Animals perform coordinated whole-body movements under the control of neural systems shaped by brain-wide connectivity. The mapping of the whole-brain neural connections, or the connectomes, provides a natural graph for modeling sensorimotor information flow, yet its potential as a neural controller for embodied agents remains largely unexplored. Here, we introduce the Fly-connectomic Graph Model, which directly instantiates the whole-brain connectome of an adult Drosophila as a graph-structured neural controller for movements of a simulated biomechanical fruit fly via deep reinforcement learning.  We achieve stable performance across diverse locomotion tasks, as well as better sample efficiency compared to both graph and non-graph baselines. Our results demonstrate a biologically informed way towards effective control policy design by translating whole-brain wiring principles into actionable architectural priors, while also improving the interpretability through dynamic information flow. This work also highlights the potential to bridge neuromechanics with embodied intelligence by providing a computational platform for investigating the sensorimotor transformation underlying animal behavior and a paradigm to advance the development of more nature-aligned intelligent systems.

\vspace{0.5em}

Project Page \url{https://lnsgroup.cc/research/FlyGM}.

\end{abstract}

\section{Introduction}

Understanding the embodied learning of movement control is a long-standing challenge shared by artificial intelligence and neuroscience. Recent advances in neuroscience have revealed whole-brain connectomics in animals at synaptic resolution~\citep{Dorkenwald2024a}. These progress enables the possibility of linking whole-brain neural networks to the control of whole-body movement.

Advances in deep reinforcement learning have produced neural network controllers that can accomplish challenging movement tasks. However, these methods typically use hand-crafted networks (e.g., multilayer perceptrons). While effective for specific tasks, such neural architectures diverge significantly from biological circuits, limiting the interpretability of the learned computations and making it difficult to relate them to real nervous systems. Connectome-constrained models have provided insights into sensory and premotor computations~\citep{Azevedo2024}, yet most efforts remain limited to specific subsystems and simplified behaviors. The fundamental challenge remains: how can static connectomes be transformed into dynamic, functional models that reproduce the intricate and adaptive motor behaviors of bodies? Answering this question requires bridging two active lines of research: (i) mechanistic modeling that leverages whole-brain connectivity, and (ii) learning frameworks that can control high-dimensional whole-body movements.

In this work, we develop the Fly-connectomic Graph Model (FlyGM), a neural network controller whose computational architecture is directly adapted from the Drosophila whole-brain connectome. We construct the control model as a directed message-passing graph, in which nodes are partitioned into afferent, intrinsic, and efferent sets according to their neurophysiological functions and edges encode real neural connectivity to preserve biologically grounded information flow. We then train this connectome-grounded policy with deep reinforcement learning to control the movements of a physics-based fruit fly model~\citep{Vaxenburg2025}.

Our experiments show that FlyGM achieves stable control across multiple locomotion tasks, including gait initiation, walking, turning, and flight. We verify the significance of this structural prior by comparing FlyGM against a degree-preserving rewired graph, a random graph, and multilayer perceptrons (MLP). Our results reveal that the connectome-structured model yields consistently lower angle error across all conditions and faster convergence during imitation learning, suggesting that the connectome's organization is non-randomly optimized for the constraints of a physical body.

Beyond behavioral performance, we analyze the internal representations of FlyGM. We find that functional segregation across sensory, central, and motor populations emerges naturally from the trained dynamics. This differentiation occurs primarily due to the structural constraints of the graph, providing evidence that biological wiring diagrams can induce functional specialization in artificial agents.

Our main contributions are summarized below:

\textbf{Connectome-Structured Architecture}: FlyGM is a graph neural controller instantiated from the whole-brain Drosophila connectome, transforming static biological wiring into a dynamic controller.

\textbf{Diverse Embodied Locomotion}: FlyGM enables whole-body biomechanical control across diverse tasks in high-fidelity physics simulations.

\textbf{Evidence of Structural Inductive Bias}: The biological connectome yields higher sample efficiency and performance compared to baseline models.

\begin{figure}[htbp]
  \centering
  \includegraphics[width=\linewidth]{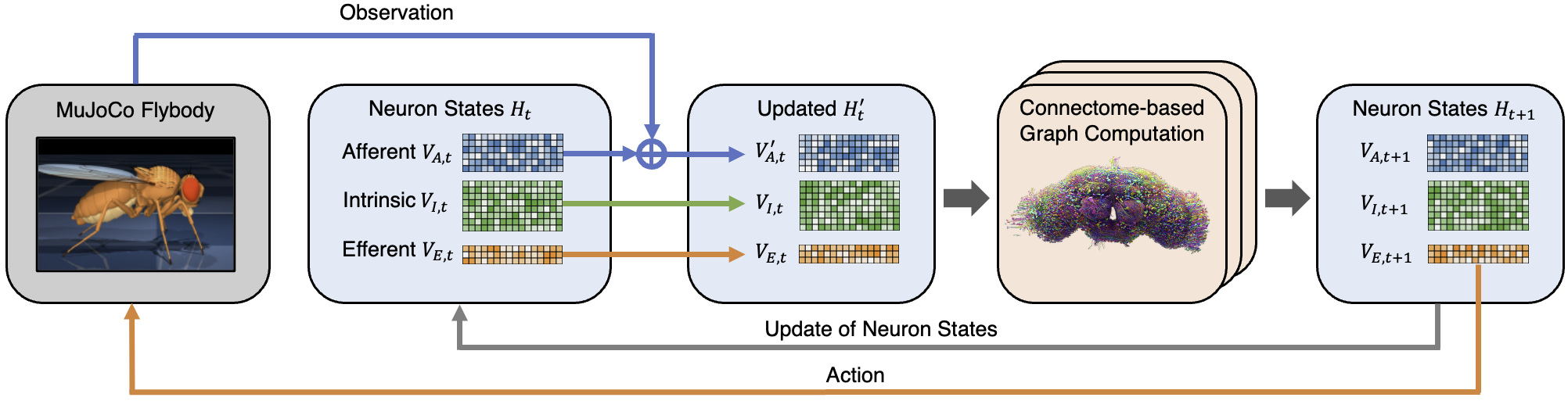}
  \caption{\textbf{Overview of the FlyGM-enabled whole-body locomotion control framework.} Observations are mapped to afferent neuron states via a lightweight input projection. Neural states are propagated through a connectome-constrained message-passing module, and the resulting efferent states are decoded into motor actions that drive the embodied Drosophila model in MuJoCo. Demonstration videos and source code are available at: \url{https://sites.google.com/view/flygm}.}
  \label{fig:mainfig}
\end{figure}


\section{Related Work}

\textbf{Connectomics-based neural network modeling.} Advances in connectomics have deepened our understanding of circuit-level organization in animal central nervous systems. In particular, the FlyWire project provides a whole-brain reconstruction of Drosophila at synaptic resolution~\citep{Dorkenwald2024a,Schlegel2024a,Lin2024a,Zheng2018a}, enabled by automated synapse detection and segmentation methods~\citep{Buhmann2021,Heinrich2018a}, offering the structural basis for modeling complete neural dynamics. Prior works have highlighted the explanatory power of such data in restricted domains: ventral nerve cord reconstructions revealed leg-wing coordination circuits~\citep{Azevedo2024,Lesser2024}, and models of motor neurons have been applied to feeding or grooming behaviors~\citep{Shiu2024}. Connectome-constrained networks have also been used to predict neural activity in the visual system~\citep{Lappalainen2024,Matsliah2024a}, and biologically constrained recurrent networks impose Dale's law and topological priors without leveraging complete connectome wiring~\citep{Balwani2025}. In other organisms, integrative data-driven models simulate brain-body-environment interactions in C.~elegans~\citep{Zhao2024}, and embodied agents have been used to recapitulate whole-brain dynamics in virtual zebrafish~\citep{keller2025}. However, these approaches often focus on specific subsystems, simpler nervous systems, or partial constraints, leaving open the question whether whole-brain connectivity can generate realistic control of embodied locomotion behaviors.

\textbf{Embodied movement control.} In parallel, embodied intelligence research has advanced locomotion in simulated humanoids~\citep{Kumar2021,Cheng2024}, quadrupeds~\citep{Ding2021}, and musculoskeletal agents~\citep{he2024icml,Wei2025,Chiappa2024} using reinforcement learning. RL-trained recurrent agents have also been shown to reproduce insect-like emergent behaviors and analyzable neural dynamics in sensorimotor tasks~\citep{Singh2023}. In the Drosophila domain, physics-based models such as NeuroMechFly~\citep{LobatoRios2022,WangChen2024} and flybody~\citep{Vaxenburg2025} have enabled detailed simulations of walking and flight in MuJoCo~\citep{todorov2012mujoco}. Yet, controllers for these systems are typically built from generic MLPs or manually designed central pattern generators, lacking direct biological grounding. This limits both interpretability and the ability to connect neural structure to behavior. Our work differs by directly embedding the connectome into the controller architecture, combining embodied simulation with structural priors to study both performance and neural representation.

\begin{figure}[htbp]
  \centering
  \includegraphics[width=\textwidth]{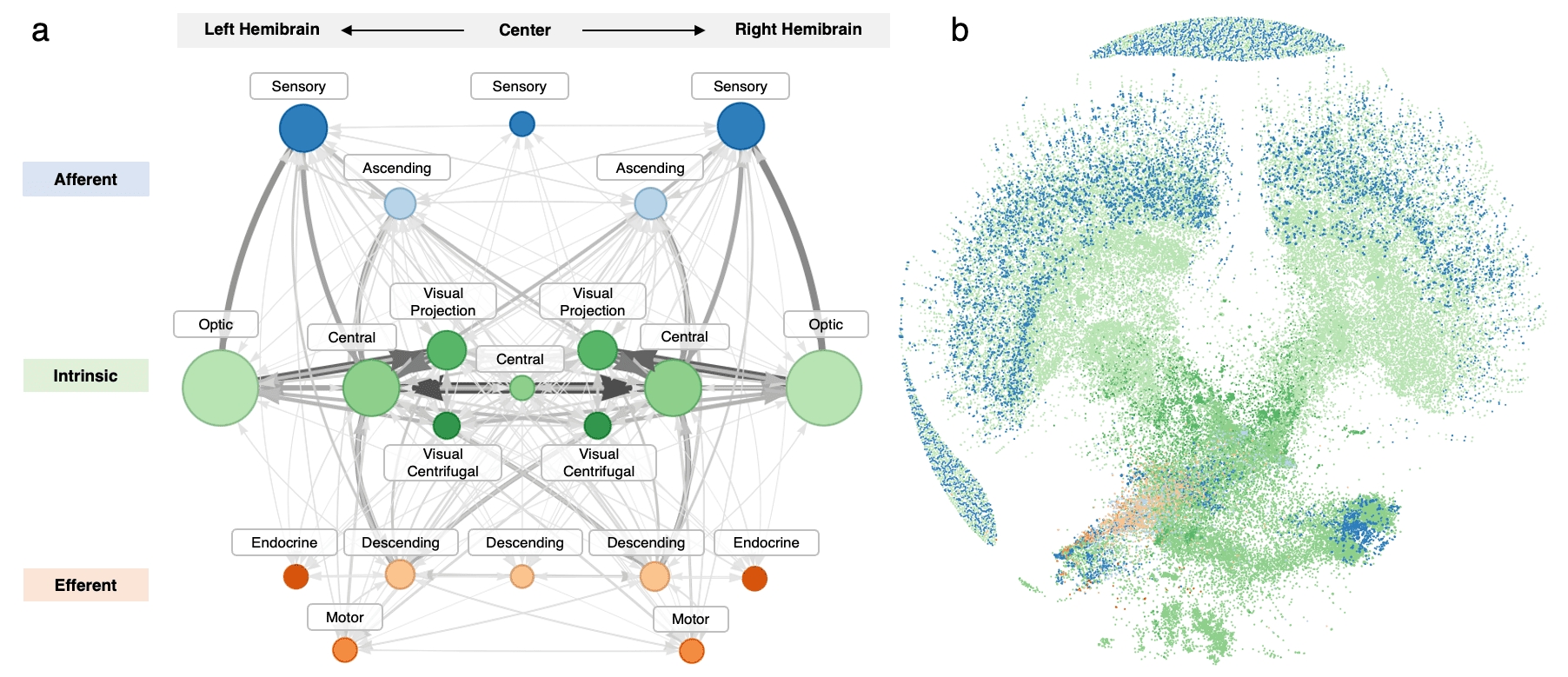}
  \caption{\textbf{Structure of the Fly-connectomic Graph.} \textbf{(a)} Aggregated synapse graph of the fly connectome, grouped into afferent (blue), intrinsic (green), and efferent (orange) sets across left hemibrain, central, and right hemibrain compartments. Node sizes reflect the number of neurons in each group, and arrows indicate the direction and relative strength of connectivity. \textbf{(b)} Force-directed graph layout~\citep{Kobourov2012} of the same neural network. The spatial layout reveals hemispheric symmetry and functional clustering.}
  \label{fig:connectome}
\end{figure}

\section{Method}
\label{sec:method}

\subsection{Connectome-structured Graph Model}

We consider the problem of embodied sensorimotor control for a virtual fruit fly agent interacting with a physics-based environment provided by flybody. Let the state at time step $t$ be denoted by $s_t \in \mathcal{S}$. The agent receives an observation $x_t \in \mathbb{R}^{d_{\text{in}}}$, which corresponds to a set of processed features including proprioceptive and exteroceptive signals during movement and environmental interaction. Based on this input, the neural controller produces an action $a_t \in \mathbb{R}^{d_{\text{out}}}$, representing motor outputs that drive the flybody model to perform locomotion behaviors.

We represent the Drosophila connectome as a directed synapse graph $G=(V,E)$, where nodes $V$ denote neurons and edges $E$ represent synaptic connections. Following the linear-dynamical connectome modeling perspective in ``effectome'' \citep{Pospisil2024}, the synaptic weight matrix $W \in \mathbb{R}^{|V| \times |V|}$ acts as a fixed, recurrent state-transition operator that dictates the information flow through the network.

To ground the model's latent dynamics in biological reality, we refine the synaptic weights by incorporating the functional polarity of presynaptic neurons. Based on established Drosophila cell type annotations and neurotransmitter predictions from electron microscopy~\citep{Eckstein2024a}, we categorize major neurotransmitters into two functional sets:

\textbf{Excitatory ($\mathcal{S}_{\text{exc}}$):} Acetylcholine (ACh), Glutamate (GLU), Aspartate (ASP), and Histamine (HIS).
\textbf{Inhibitory ($\mathcal{S}_{\text{inh}}$):} $\gamma$-aminobutyric acid (GABA) and Glycine (GLY).

Following this categorization, the signed strength $W_{vu}$ for a directed edge from neuron $u$ to $v$ is defined as the net polarized synaptic count:
\begin{equation}
    W_{vu} = N_{\text{exc}}(u, v) - N_{\text{inh}}(u, v)
\end{equation}
where $N_{\text{exc}}$ and $N_{\text{inh}}$ are the total synapse counts associated with excitatory and inhibitory neurotransmitters, respectively. This signed weighting scheme ensures that the message-passing operation $M_t = W H_t$ respects the sign-preserving or sign-reversing nature of biological signal propagation, consistent with the methodology proposed by \citet{Pospisil2024}.

Following FlyWire's classification of flow types, we partition neurons into three disjoint sets:
\begin{itemize}
\setlength{\itemsep}{0pt}
\setlength{\parsep}{0pt}
\setlength{\parskip}{0pt}
\item \textbf{Afferent neurons}: $V_a \subset V$, which receive external sensory inputs,
\item \textbf{Intrinsic neurons}: $V_i \subset V$, which mediate signals within the network, and
\item \textbf{Efferent neurons}: $V_e \subset V$, which produce motor outputs to the body model.
\end{itemize}

Thus, $V = V_a \cup V_i \cup V_e$ with $V_a \cap V_i = V_a \cap V_e = V_i \cap V_e = \emptyset$. Each neuron $v \in V$ is associated with a latent state vector $h_{v,t} \in \mathbb{R}^{C}$, and we collect all neuron states at time $t$ as a matrix:
\begin{equation}
H_t = \begin{bmatrix} h_{1,t} & \cdots & h_{|V|,t} \end{bmatrix}^\top \in \mathbb{R}^{|V| \times C}.
\end{equation}

Beyond the shared synaptic operator $W$, we assign each neuron a trainable intrinsic descriptor $\eta_v \in \mathbb{R}^{D}$ to capture cell-specific computational properties (e.g., excitability and gain) that are not included in the connectomic data. The matrix stacking all intrinsic descriptors is represented as $\eta \in \mathbb{R}^{|V|\times D}$.

At each time step, sensory observation $x_t$ is first encoded by an encoder $\mathrm{Enc}_\theta$ and injected into afferent neurons via a gating map:
\begin{align}
\tilde{x}_t &= \mathrm{Enc}_\theta(x_t) \in \mathbb{R}^{d_{\text{enc}}},\\
H_t[V_a] &\leftarrow \tanh\!\big(W_g\,[\,H_t[V_a] \,\Vert\, \mathbf{1}\tilde{x}_t^\top\,] + b_g\big),
\end{align}
where $\mathbf{1}\tilde{x}_t^\top$ propagates the encoded observation to all afferent neurons. The encoder $\mathrm{Enc}_\theta$ compresses the high-dimensional observation $x_t$ into a low-dimensional representation (e.g., $d_{\text{enc}}=32$) before projecting to $V_a$, avoiding expensive per-neuron encoding and allowing the connectome-conditioned recurrent dynamics to carry out most of the computation.

Given the current network state $H_t$, we compute synaptic aggregation by applying connectome-derived synaptic weights as a linear operator, analogous to message passing in graph neural networks~\citep{Kipf2017a,Hamilton2018,Corso2020}:
\begin{equation}
M_t = W H_t \in \mathbb{R}^{|V|\times C},
\end{equation}
so that each neuron receives the weighted sum of its presynaptic partners' states.

We then update each neuron state through a shared MLP $f_\psi$ conditioned on the neuron's intrinsic descriptor:
\begin{equation}
H_{t+1}[v] = f_\psi\big(\,[\,M_t[v] \,\Vert\, \eta_v\,] \,\big), \quad v\in V.
\end{equation}
The updated efferent states $H_{t+1}[V_e]$ are flattened and mapped to continuous motor actions by a decoder $\mathrm{Dec}_\phi$:
\begin{equation}
a_t = \mathrm{Dec}_\phi\!\big(H_{t+1}[V_e]\big).
\end{equation}
The output actions serve as motor commands to actuate the flybody, a biomechanical model of the fruit fly implemented in MuJoCo~\citep{Vaxenburg2025}. The subsequent observation $x_{t+1}$ is then generated to close the sensorimotor loop.

In summary, Algorithm~\ref{alg:FlyGM} outlines the forward computation of FlyGM: observation injection into afferent neurons, synapse-weighted aggregation via the connectome operator, neuron-wise conditional updates using intrinsic descriptors, and efferent decoding into motor actions.

\begin{algorithm}[tb]
  \caption{Fly-connectomic Graph Model (FlyGM)}
  \label{alg:FlyGM}
  \begin{algorithmic}[1]
    \STATE {\bfseries Input:} Sensory input $x_t$; connectome $G=(V,E)$ with synapse weight matrix $W$; node partitions $V_a, V_i, V_e$;
    encoder $\mathrm{Enc}_\theta$; afferent gate $(W_g,b_g)$; neuron intrinsic descriptors $\{\eta_v\}_{v\in V}$;
    conditional update MLP $f_\psi$; decoder $\mathrm{Dec}_\phi$
    \STATE {\bfseries Output:} Motor output $a_t$

    \FOR{each time step $t$}
      \STATE $\tilde{x}_t \gets \mathrm{Enc_\theta}(x_t)$
      \STATE $H_t[V_a] \gets \tanh\!\big(W_g [H_t[V_a], \mathbf{1}\tilde{x}_t^\top] + b_g\big)$
      \STATE $M_t \gets W H_t$
      \STATE $H_{t+1}[v] \gets f_\psi([M_t[v], \eta_v]) \quad \forall v\in V$
      \STATE $a_t \gets \mathrm{Dec_\phi}(H_{t+1}[V_e])$
      \STATE Apply $a_t$ to flybody in MuJoCo to obtain $x_{t+1}$
    \ENDFOR
  \end{algorithmic}
\end{algorithm}

\subsection{Training Pipeline}

Our training pipeline consists of two stages: we first initialize the connectome-based policy using imitation learning from expert trajectories, and then fine-tune the model with reinforcement learning to directly optimize for task rewards. This two-stage design leverages demonstration data for rapid initialization while preserving the capability for adaptive policy improvement.

To provide an initial policy, we collect expert trajectories by rolling out an MLP-based policy for the flybody which was originally trained with imitation learning to generate high-quality demonstrations of locomotion. We then use these trajectories to train our connectome-based model by imitating the expert's action distributions.

Specifically, the policy predicts Gaussian parameters $(\mu_t, \sigma_t)$ given the same observations as the expert, and is optimized to minimize a loss combining Kullback-Leibler divergence with an annealed mean squared error (MSE) regularizer:
\begin{equation}
\begin{aligned}
    \mathcal{L}_t &= D_{\mathrm{KL}}\!\left( \mathcal{N}(\mu_t, \sigma_t^2) \;\|\; \mathcal{N}(\mu_s, \sigma_s^2) \right) + \lambda(t) \Big( \|\mu_s - \mu_t\|_2^2 + \alpha \,\|\log \sigma_s - \log \sigma_t\|_2^2 \Big)
\end{aligned}
\end{equation}
where $(\mu_s, \sigma_s)$ are the Gaussian parameters predicted by the expert (source) policy on the same observation, $\alpha$ is a constant balancing the scale of $\mu_s$ and $\log \sigma_s$, and $\lambda(t)$ decays during training so that distributional matching dominates in later stages. This procedure initializes the model with stable behaviors for walking and flight tasks.

After initialization, we fine-tune the connectome-structured policy using Proximal Policy Optimization (PPO) to enable direct learning from rewards. For value estimation, we use a simple MLP as the value network. The environments in MuJoCo are adapted into gym-like interfaces with parallel rollouts to increase throughput, and distributed training with Distributed Data Parallel (DDP) is used for scalability. The policy is updated following the clipped surrogate objective with value and entropy regularization:
\begin{equation}
\begin{aligned}
\mathcal{L}_{\text{PPO}} &= \mathbb{E}_t \Big[ \min\big(r_t(\theta) \hat{A}_t, \text{clip}(r_t(\theta), 1\pm\epsilon)\hat{A}_t\big) - c_v (V_\theta(s_t) - R_t)^2 + c_e \mathcal{H}[\pi_\theta(\cdot|s_t)] \Big]
\end{aligned}
\end{equation}
where $ r_t(\theta)=\frac{\pi_\theta(a_t \mid s_t)}{\pi_{\theta_{\text{old}}}(a_t \mid s_t)} $ is the probability ratio between new and old policies, $\hat{A}_t$ is the GAE advantage, $R_t$ is the return, and $\mathcal{H}$ is the entropy bonus. This stage allows the model to improve beyond demonstration data and adapt to task-specific dynamics while retaining the inductive bias imposed by the connectome architecture.

\section{Experiments}

We evaluated FlyGM on four locomotor tasks with the flybody physics simulator~\citep{Vaxenburg2025}, comprising three terrestrial tasks (gait initiation, straight walking, and turning) and one flight task. These tasks are categorized into two primary behavioral domains (walking and flying) for which flybody provided distinct pre-trained MLP controllers and real-world Drosophila motion capture datasets. We followed the default flybody environment setup and added binocular visual signals to the original sensory inputs for walking tasks to provide multimodal feedback (proprioception, mechanosensation, and vision).

To support the imitation learning stage, we constructed a domain-specific behavioral dataset by performing rollouts of the corresponding pretrained models across their respective biological trajectories. We filtered these rollouts to include only successful episodes exceeding 100 steps, resulting in a comprehensive dataset of synchronized observation-action time series. This curated dataset captures the high-fidelity, naturalistic motor patterns necessary to test whether a connectome-structured network can flexibly generate stable control policies across distinct movement modes.

Detailed training procedures and hyperparameters are provided in Appendix~\ref{app:train_details}, and observation/action definitions are provided in Appendix~\ref{app:env_details}. Demonstration videos for each task are available on our project webpage.

\subsection{Inductive Bias from Connectome Topology}
\label{sec:inductive_bias}

To evaluate whether the biological connectome provides a structural inductive bias beyond mere parameter count, we compared FlyGM against three baselines under an identical IL+PPO pipeline. The first was a \textbf{degree-preserving rewiring} of the connectome that randomly shuffled edges while preserving each neuron's in-degree and out-degree. The second was an \textbf{Erd\H{o}s-R\'enyi random graph} with matched node and edge counts. The third was an \textbf{MLP} with four 512-unit hidden layers, whose parameter count exceeded FlyGM. For graph-based non-connectome models, all edges were unweighted with unit strength, since trainable intrinsic descriptors and injective sum aggregation already capture structural properties without additional weighting~\citep{Xu2019}. All graph models shared a latent dimension of 32. As a separate biologically inspired but non-connectome control, we also evaluated a spiking neural network (SNN) baseline; full implementation and results are reported in Appendix~\ref{app:snn}. We further ablated FlyGM's edge weights and intrinsic descriptors, and defer the full results to Appendix~\ref{app:flygm_ablation}.

\begin{figure}[htbp]
  \centering
  \includegraphics[width=\textwidth]{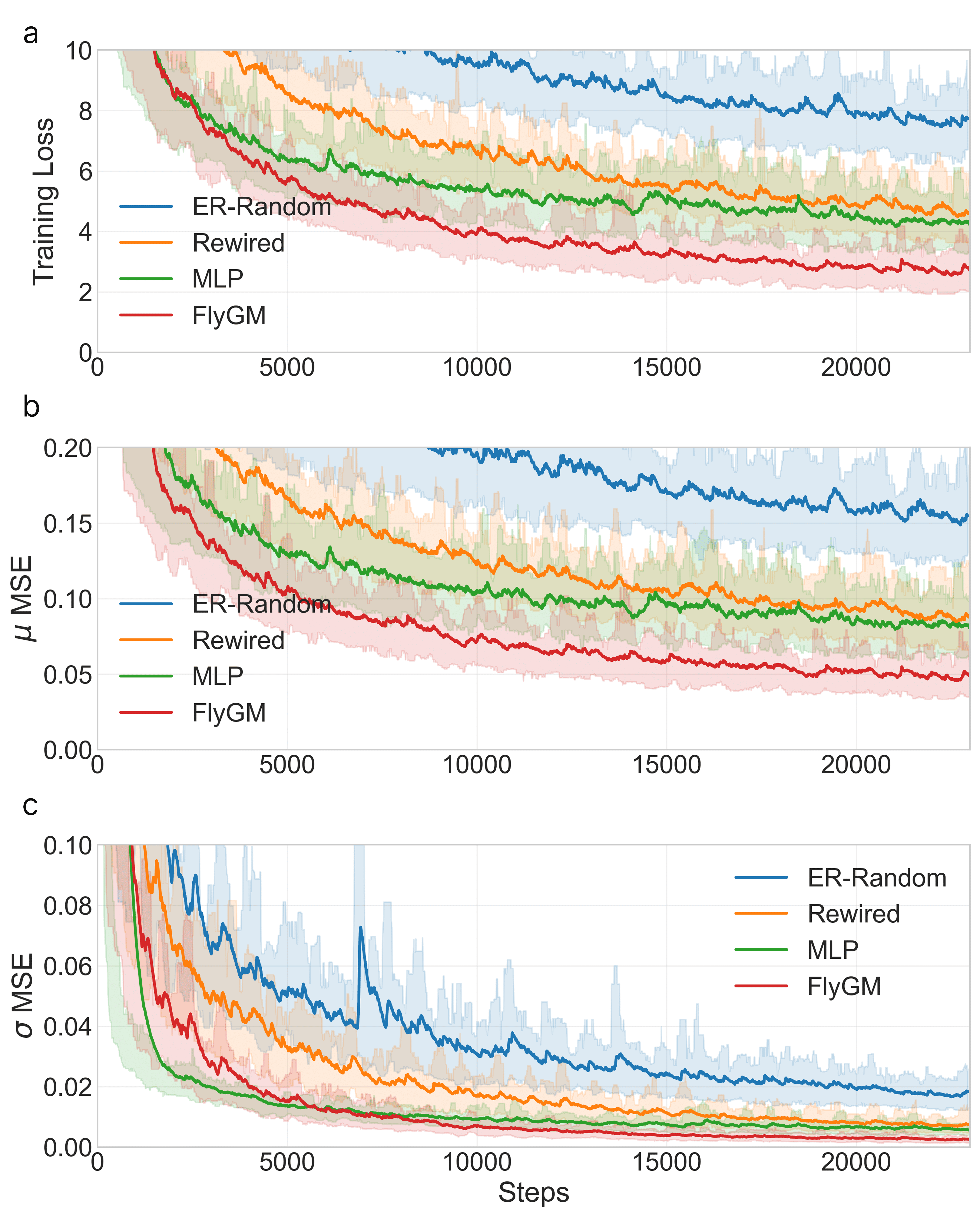}
  \caption{\textbf{Imitation learning dynamics across topologies.} (a) Total training loss and (b) mean squared error of action means ($\mu$ MSE) relative to expert trajectories. Shaded areas denote standard deviation across multiple runs. ER-Random: Erd\H{o}s-R\'enyi random graph; Rewired: degree-preserving rewired graph; MLP: multilayer perceptron.}
  \label{fig:bias}
\end{figure}

\textbf{Sample efficiency during imitation.} Figure~\ref{fig:bias} compares the imitation learning stage across topologies. The connectome-based FlyGM converges faster and reaches a lower training loss and action-mean MSE than every non-connectome baseline, indicating that the biological wiring matches expert trajectories with fewer gradient updates. The rewired and random graphs lag throughout training, and the MLP plateaus at a higher error despite its larger parameter count.

\textbf{Final control performance.} Table~\ref{tab:topology_reduction} reports control performance after the full IL+PPO pipeline across speeds $v$ (cm/s) and yaw rates $\psi$ (rad/s). The structural advantage of the connectome grows with task complexity. The degree-preserving rewiring loses orientation control as task difficulty increases, with angle error rising to $13.55$ at high yaw versus $8.29$ for FlyGM. The MLP, despite its larger parameter count, trails FlyGM on angle error in every condition. The Erd\H{o}s-R\'enyi graph degenerates catastrophically at high yaw ($125.36$). The SNN baseline (Appendix~\ref{app:snn}) fails to learn a usable gait, indicating that single-neuron biological plausibility alone cannot substitute for connectome-level structural priors. Combined with the ablations in Appendix~\ref{app:flygm_ablation}, where even an unweighted FlyGM outperforms all non-connectome baselines, these results demonstrate that FlyGM achieves the lowest angle error across all speed and yaw conditions, and that the wiring specificity of the biological connectome is what drives stable, high-fidelity locomotion control.

\begin{table}[htbp]
\centering
\small
\setlength{\tabcolsep}{6pt}
\renewcommand{\arraystretch}{1.1}
\begin{tabular}{ll cccc}
\toprule
\textbf{Model} & \textbf{Metric} & $v{=}2,\psi{=}0$ & $v{=}3,\psi{=}0$ & $v{=}3,\psi{=}4$ & $v{=}3,\psi{=}7$ \\
\midrule
{\textbf{FlyGM (ours)}}
 & Angle Err $\downarrow$ & $\mathbf{4.96{\pm}0.09}$ & $\mathbf{5.57{\pm}0.06}$ & $\mathbf{6.36{\pm}0.33}$ & $\mathbf{8.29{\pm}0.21}$ \\
\midrule
{\textbf{DP Rewiring}}
 & Angle Err $\downarrow$ & $7.84{\pm}0.08$ & $7.77{\pm}0.08$ & $9.72{\pm}0.12$ & $13.55{\pm}0.69$ \\
\midrule
{\textbf{Erd\H{o}s-R\'enyi}}
 & Angle Err $\downarrow$ & $12.11{\pm}0.23$ & $11.33{\pm}0.27$ & $17.45{\pm}0.26$ & $125.36{\pm}8.96$ \\
\midrule
{\textbf{MLP}}
 & Angle Err $\downarrow$ & $6.76{\pm}0.13$ & $7.18{\pm}0.05$ & $8.85{\pm}0.24$ & $13.90{\pm}0.45$ \\
\bottomrule
\end{tabular}
\vspace{1em}
\caption{Control performance across speed ($v$, cm/s) and yaw ($\psi$, rad/s) conditions after the full IL+PPO pipeline. Angle error in degrees. Values are mean$\pm$std with the per-column best mean in bold. DP Rewiring denotes the degree-preserving rewiring of the connectome. SNN baseline results are reported separately in Appendix~\ref{app:snn}.}
\label{tab:topology_reduction}
\end{table}

\subsection{Performance on Locomotion Tasks}

\begin{figure}[htbp]
  \centering
  \begin{subfigure}{0.4\textwidth}
    \centering
    \includegraphics[width=\textwidth]{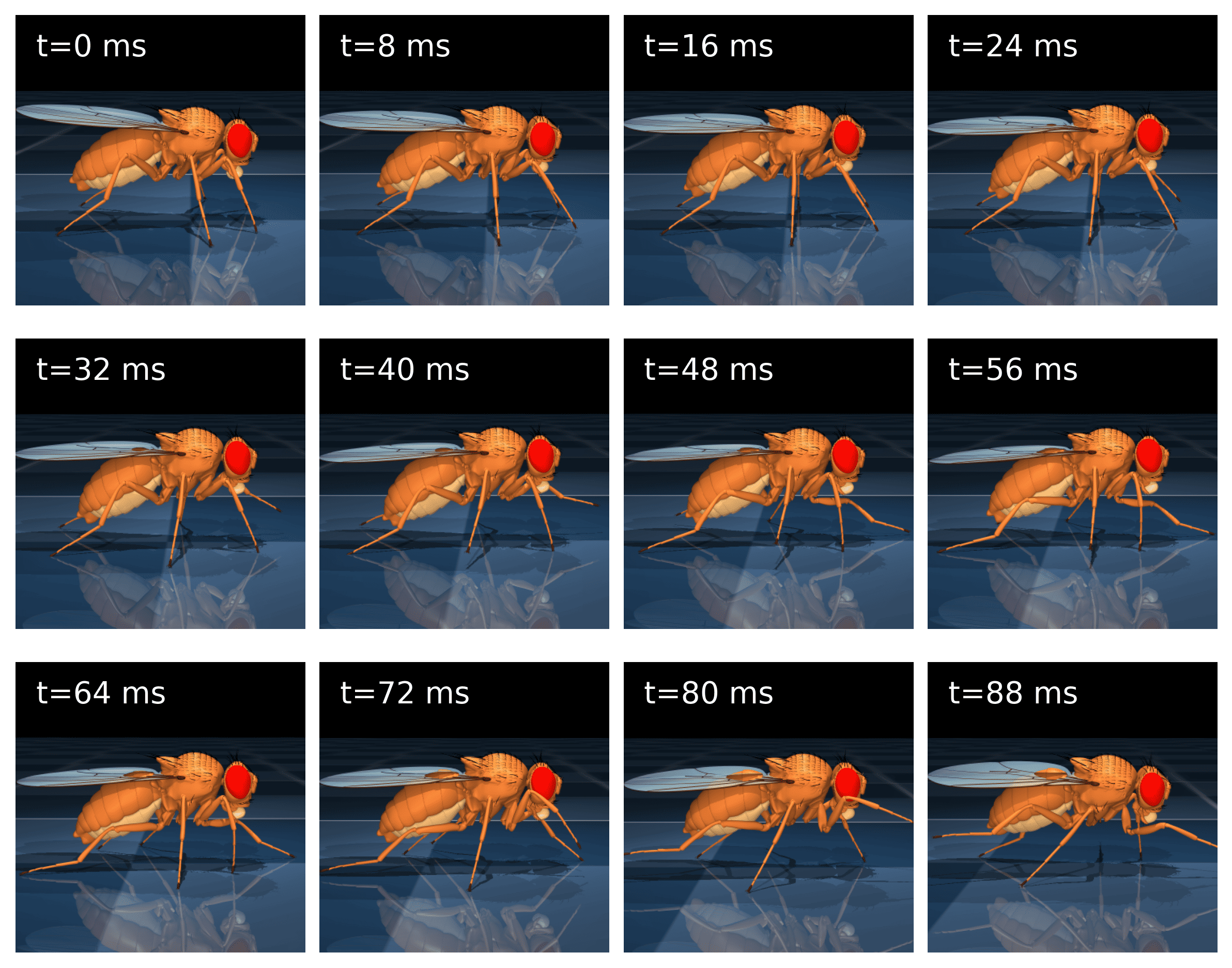}
    \caption{Gait initiation dynamics.}
    \label{fig:flystart}
  \end{subfigure}
  \hspace{5em}
  \begin{subfigure}{0.4\textwidth}
    \centering
    \includegraphics[width=\textwidth]{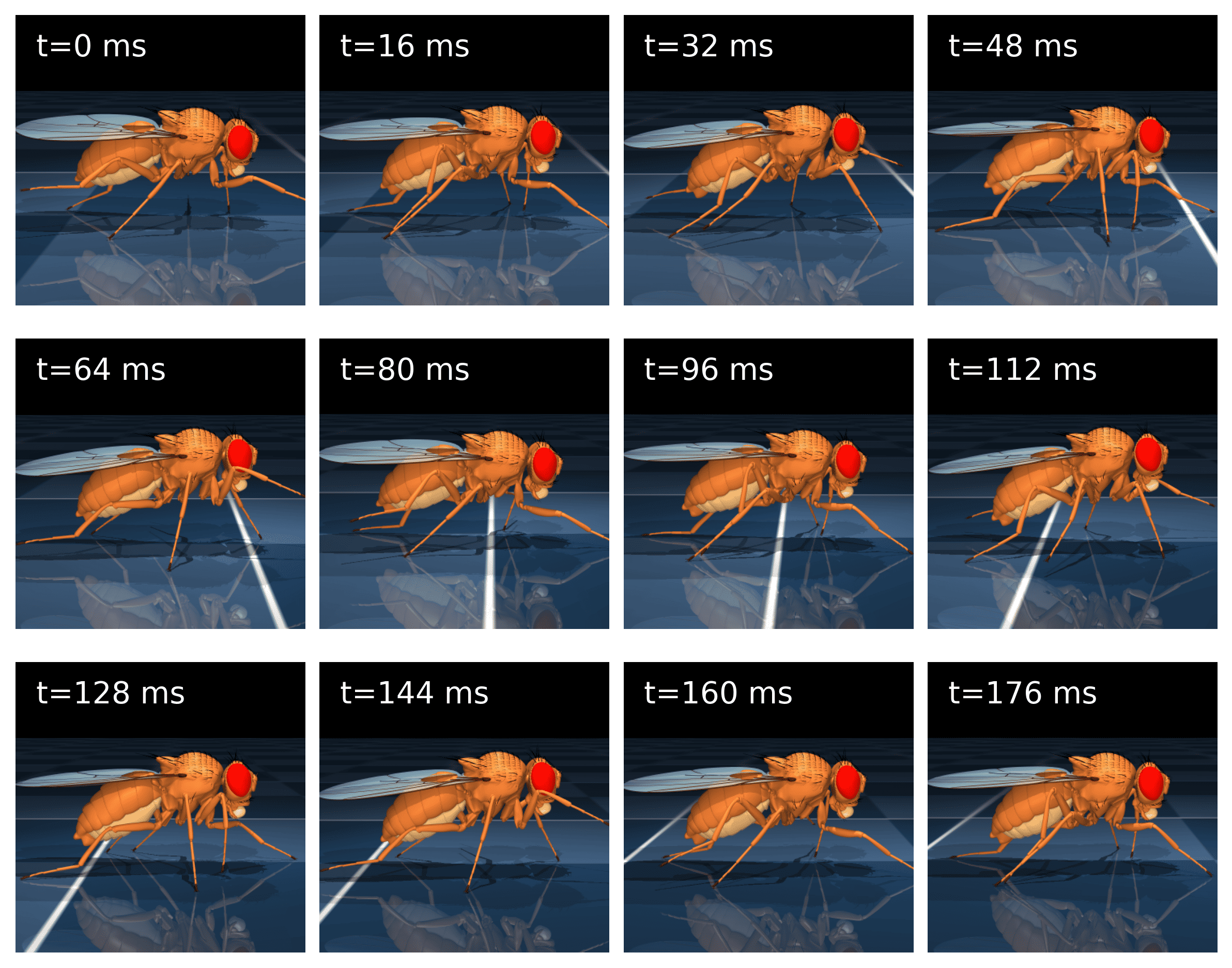}
    \caption{Walking dynamics.}
    \label{fig:flywalk}
  \end{subfigure}
  \caption{\textbf{Walking task snapshots.} (a) Snapshots of the simulated fly during the gait initialization phase. (b) Snapshots of the simulated fly walking in a straight line at a velocity of 3 cm/s.}
  \label{fig:walk_snapshots}
\end{figure}

\textbf{Gait Initiation.}

Building on the training pipeline described above, we evaluated FlyGM on a walking task at a target velocity of 3 cm/s. We first examined the process of gait initiation, focusing on the transition from rest to the onset of stable locomotion. Figure~\ref{fig:flystart} illustrates snapshots of the simulated fly prior to the first complete gait cycle. It successfully initiated the movement in roughly the first 80 ms, during which irregular steps gradually gave way to rhythmic and coordinated leg movements.

\textbf{Straight-Line Walking.}

Following the analysis of gait initiation, we next evaluated FlyGM on the straight-line walking phase at a target velocity of 3 cm/s. As illustrated in Figure \ref{fig:flywalk}, the flybody model performed stable forward locomotion with clear tripod coordination. The simulated fly maintained a consistent body trajectory over hundreds of milliseconds without drift or collapse, indicating that the learned controller generalized well to continuous walking.

Joint-level analysis (Figure~\ref{fig:coxa} in Appendix~\ref{app:kinematics}) shows that actuator outputs are tightly coupled with kinematic trajectories: contralateral legs alternate in phase, producing the classical tripod gait pattern seen in Drosophila. These results demonstrate that FlyGM is sufficient to generate stable straight walking once locomotion is initiated.

\begin{figure}[htbp]
  \centering
  \begin{subfigure}{0.4\textwidth}
    \centering
    \includegraphics[width=\textwidth]{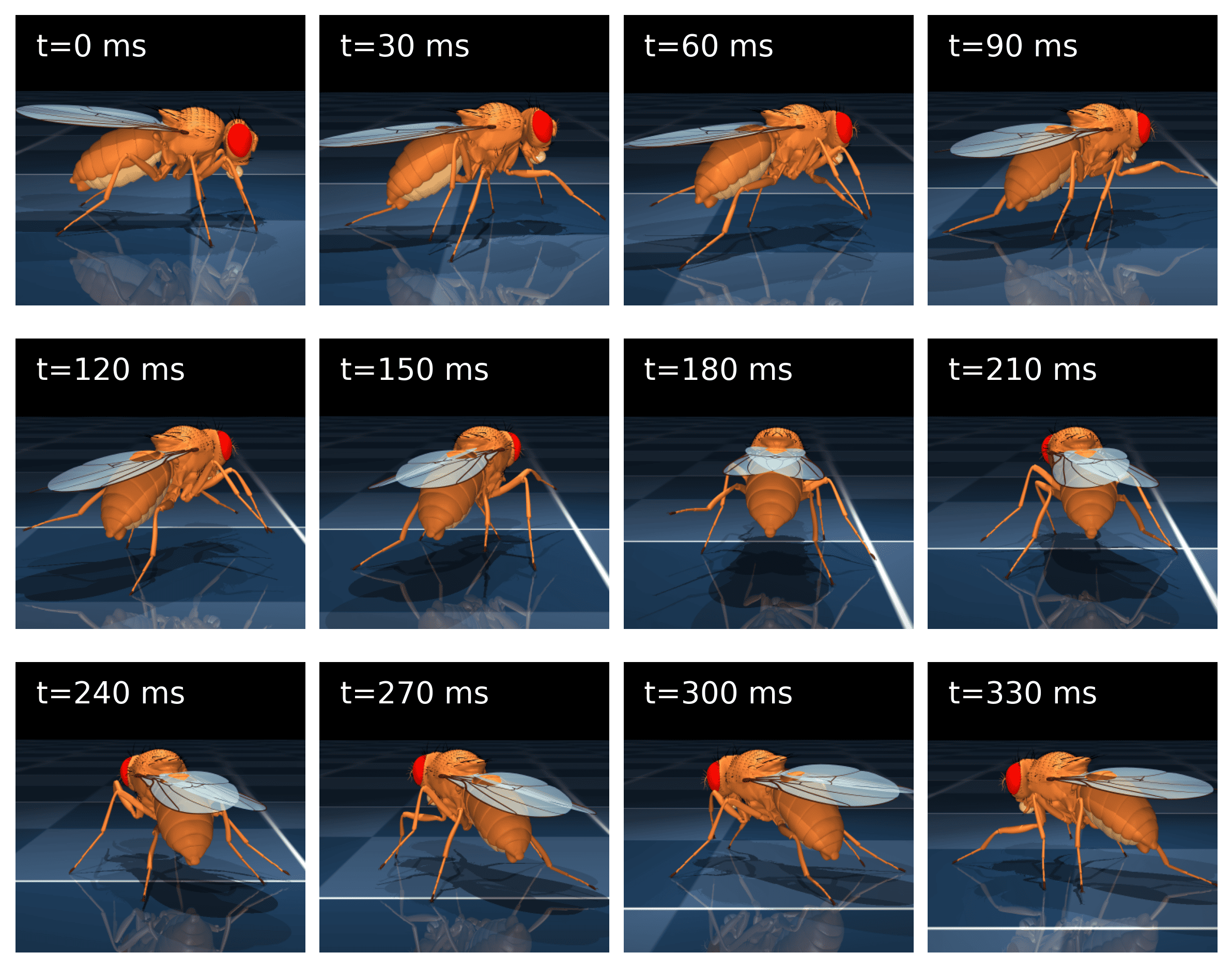}
    \caption{Turning dynamics.}
    \label{fig:flyturn}
  \end{subfigure}
  \hspace{5em}
  \begin{subfigure}{0.4\textwidth}
    \centering
    \includegraphics[width=\textwidth]{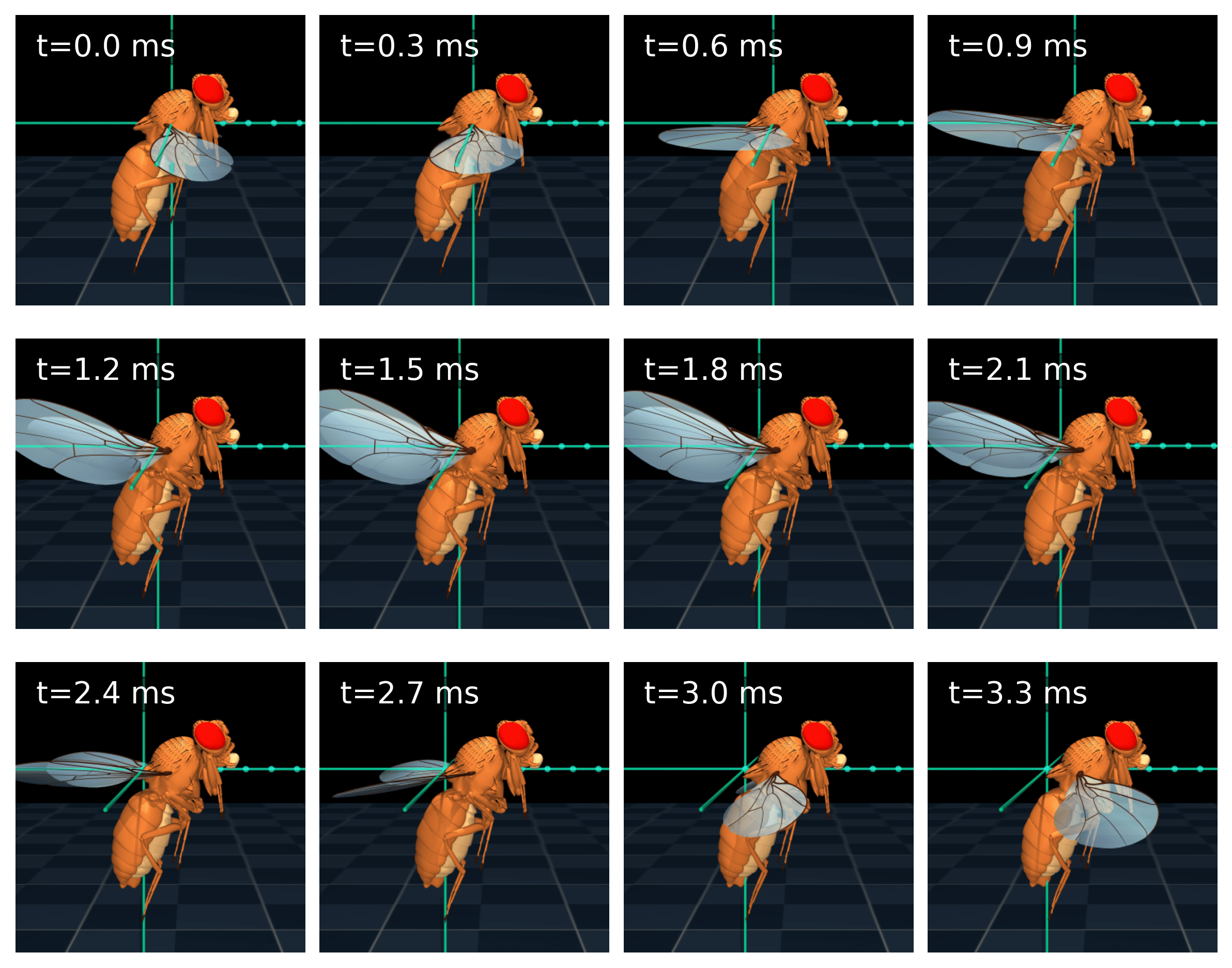}
    \caption{Flight dynamics.}
    \label{fig:flyflight}
  \end{subfigure}
  \caption{\textbf{Turning and flight task snapshots.} (a) Snapshots of the virtual fly executing a high-speed left turn at 3 cm/s and 10 rad/s. (b) Snapshots of the virtual fly executing a straight flight task at a velocity of 20 cm/s.}
  \label{fig:turn_flight_snapshots}
\end{figure}

\textbf{Turning.}

We next assessed whether the same policy could generalize to directional maneuvers. In the turning task, the model was instructed to walk at a forward velocity of 3 cm/s while executing a leftward turn at 10 rad/s. As shown in Figure \ref{fig:flyturn}, the simulated fly successfully produced a smooth curved trajectory by modulating stride lengths asymmetrically across the body: legs on the inner side of the turn reduced their stance amplitude, while contralateral legs extended their strides. This modulation of gait symmetry arose naturally from the network dynamics, without requiring task-specific tuning or additional control rules.

\textbf{Flight.}

To assess whether FlyGM can generalize beyond terrestrial locomotion, we additionally trained FlyGM to perform a flight task. In this setting, the policy served as a higher-level neural controller that modulated the output of the wing-beat pattern generator, thereby enabling stable flight dynamics.

As shown in Figure~\ref{fig:flyflight}, the trained controller maintained stable forward flight at a constant speed and kept body orientation aligned with the target direction, demonstrating that the connectome-based network can extend from walking to flight locomotion.

\subsection{Neural Representation Analysis}

\begin{figure}[htbp]
  \centering
  \includegraphics[width=0.98\textwidth]{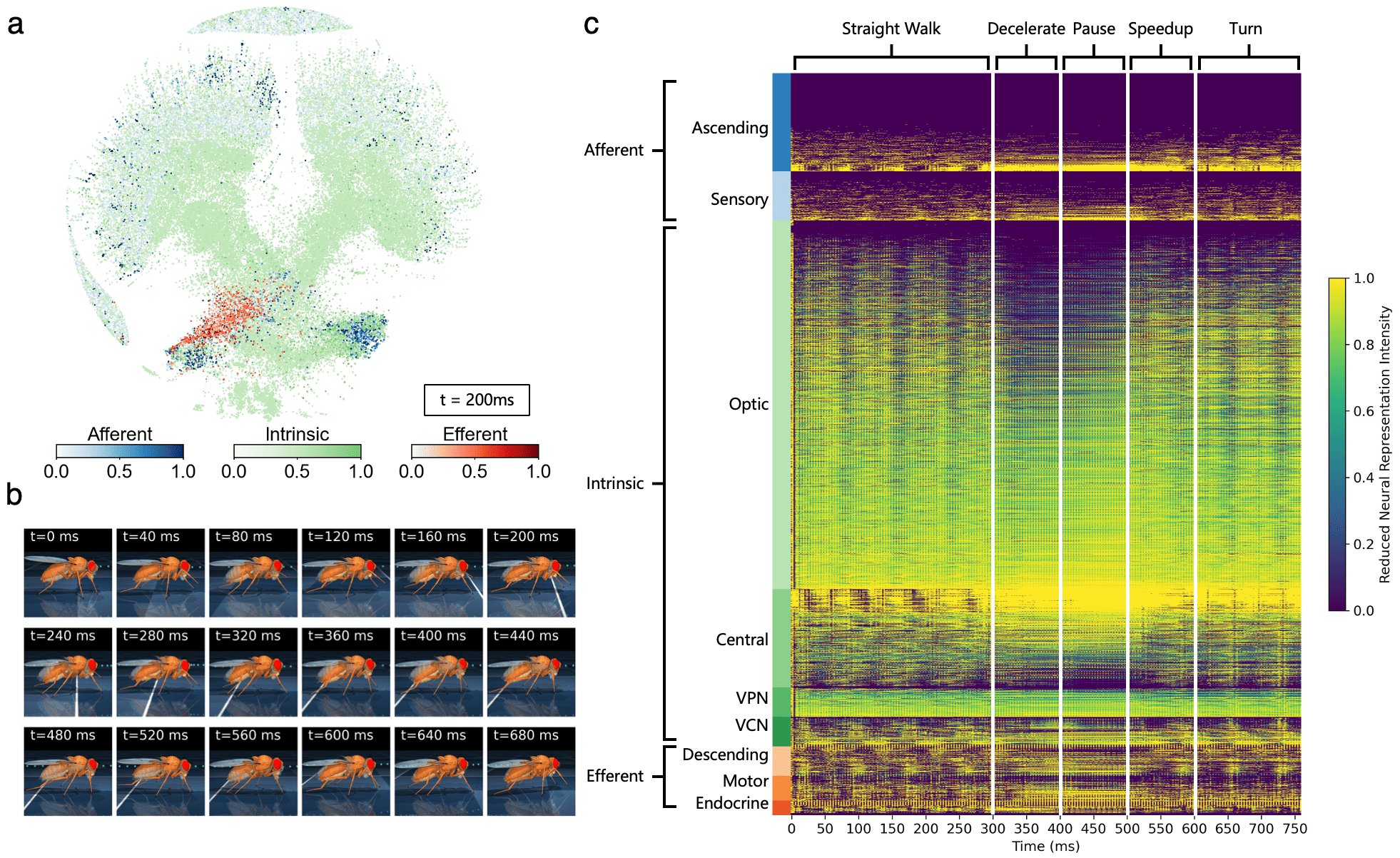}
  \caption{\textbf{Neural and behavioral dynamics across a continuous walking sequence.}
  \textbf{(a)} Force-directed layout of the connectome at 200 ms, with neurons colored by flow type and representation intensity.
  \textbf{(b)} Behavioral snapshots (0-680 ms) show straight walking, decelerating, pausing, speeding-up, and turning phases; blue dots and ghost fly mark target trajectory and orientation.
  \textbf{(c)} Temporal dynamics across the connectome, grouped by flow type and major neuron classes and showing transitions between behavioral phases.
  }
  \label{fig:repr_v2}
\end{figure}

Connectome-based architectures provide a unique opportunity to study how information propagates through biological wiring. By recording neuron states during simulation, we use the \textbf{reduced neural representation intensity} as a scalar proxy for the engagement level of individual neurons, mapping complex multi-channel latent dynamics onto a single-intensity scale that facilitates the visualization of population-level information flow (see Appendix~\ref{app:repr_analysis} for the full computation, downsampling, and neuron-ordering procedure).

Figure~\ref{fig:repr_v2} presents the temporal dynamics of this intensity over a long composite locomotion sequence. Figure~\ref{fig:repr_v2}a visualizes the connectome at 200 ms using a force-directed layout that spatially arranges neurons based on their connectivity, with neurons colored by flow type and shaded by representation intensity, revealing the activation patterns of different flow types at a specific moment. Figure~\ref{fig:repr_v2}c shows the temporal dynamics across the connectome, where neurons are grouped by superclass and reordered within each class so that those with similar activation patterns are positioned adjacently. The resulting plots demonstrate heterogeneous responses to phase transitions across different neuron superclasses, whereas neurons within the same superclass display relatively consistent response patterns. Distinct activation patterns emerge for each class, and transitions between behavioral phases imply how sensory observations and motor intentions are dynamically transmitted through the network. Such patterns are consistent with experimental neuroscience findings~\citep{Brezovec2024,schaffer2023spatial}, showing the potential to align FlyGM dynamics with neurophysiological processes. As a control, applying the same analysis to a model trained on an Erd\H{o}s-R\'enyi random graph (Appendix~\ref{app:repr_rand}, Figure~\ref{fig:repr_rand}) yields no comparable class-wise differentiation, indicating that the emergent functional segregation in FlyGM is driven by connectome-specific wiring rather than the analysis pipeline itself.

\section{Conclusion and Discussion}

This work demonstrates that the whole-brain connectome-based graph model FlyGM can serve as an efficient reinforcement learning controller to achieve diverse locomotion tasks of a simulated biomechanical fruit fly. By structuring information flow according to the FlyWire connectome, FlyGM replaces hand-crafted policy networks with a graph-based network that directly reflects the wiring diagram of the brain.

Our results demonstrate that structural priors at the connectome scale provide a powerful inductive bias for embodied control. Even when simplified to an unweighted directed graph without synapse counts or neurotransmitter types, the connectome is sufficient to drive diverse high-dimensional motor control. This suggests that wiring diagrams, long viewed as static anatomical maps, can be directly instantiated as functional networks for closed-loop control.

From the perspective of machine learning, FlyGM offers an anchor point for neural architecture search. Instead of relying on existing artificial structures, we can adapt structures from evolved biological networks. This provides a systematic alternative to generic architectures, potentially improving explainability, data efficiency, stability, and generalizability across tasks. The approach also opens the possibility of scaling to larger connectomes and more complex embodied agents, where the inductive biases of real nervous systems may be particularly advantageous.

We identify several limitations and directions for future work. First, more detailed and complete connectome data, such as updated synapse detections~\citep{Yu2025} and refined cell-type or sex-specific annotations~\citep{Matsliah2023,Deutsch2025}, could be implemented to improve our model. Incorporating richer information may improve biological fidelity. Second, compared to MLP-based controllers, our model requires longer per-step computation and higher memory usage. Finally, extending the framework beyond locomotion will provide a more comprehensive test of the generality of connectome-based control.

In summary, FlyGM transforms static whole-brain connectomes into a dynamic model that can be used for whole-body locomotion control. By grounding policy architectures in biological wiring diagrams, this approach suggests a direction towards more nature-aligned AI systems, and provides a computational platform for understanding sensorimotor control.

\newpage
\bibliography{refs}
\bibliographystyle{mybst}

\newpage
\appendix

\section{Details of Model Architecture and Training Process}
\label{app:train_details}

\subsection{Model Architecture and Implementation}

The FlyGM model instantiates the Drosophila connectome as a large-scale recurrent neural system: synapse weights define a sparse linear aggregation operator, and each neuron additionally carries trainable intrinsic parameters that condition how synaptic input updates its state. The network is structured as a directed graph where nodes represent neurons and edges represent synaptic connections, partitioned into afferent, intrinsic, and efferent sets based on information flow. The model processes inputs through the following components:

Input Normalization: A RunningNorm layer is applied to observations to stabilize training by maintaining running estimates of mean and variance. This layer operates online and is updated during training.

Encoder: Observations are projected into the afferent neuron state space via a linear layer followed by ReLU activation.

Graph Propagation: Information is propagated through the connectome using a message-passing mechanism with gated updates combining previous states and incoming messages.

Decoder: Efferent neuron states are aggregated and passed through a multi-layer perceptron with ReLU activations, outputting action means and standard deviations via linear and softplus heads, respectively.

All experiments were conducted on servers equipped with NVIDIA A100 80GB PCIe GPUs and Intel Xeon Gold 6348 CPUs.

\subsection{Training Pipeline}

Training proceeds in two stages:

Imitation Learning: The policy is initialized by mimicking expert trajectories generated by a pre-trained MLP controller provided by flybody. This stage uses PyTorch Lightning for distributed training across multiple workers.

Reinforcement Learning: The model is fine-tuned with Proximal Policy Optimization to maximize task rewards. This stage is implemented using PyTorch Distributed Data Parallel (DDP) for scalability.

Optimization uses AdamW with a learning rate scheduler (ReduceLROnPlateau) that reduces the learning rate when validation loss stops improving. Both stages employ gradient clipping to ensure stability.

\subsection{Task-Specific Configurations}
Walking Task:
\begin{itemize}
\setlength{\itemsep}{0pt}
\setlength{\parsep}{0pt}
\setlength{\parskip}{0pt}
    \item Observation dimension: 1,253 (proprioceptive + exteroceptive + visual inputs),
    \item Action dimension: 59 (joint actuators and adhesion controls),
    \item Learning rate: $1 \times 10^{-4}$,
    \item Message-Passing: 32 node channels, 4 message-passing layers,
    \item Expert policy architecture: 512-512-512-512 fully-connected layers with LayerNorm + Tanh (first layer) and ELU activations, outputting mean (Linear) and std (Linear + Softplus) for 59-dimensional actions.
\end{itemize}

Flight Task:
\begin{itemize}
\setlength{\itemsep}{0pt}
\setlength{\parsep}{0pt}
\setlength{\parskip}{0pt}
    \item Observation dimension: 104 (simplified proprioceptive and kinematic inputs),
    \item Action dimension: 12 (wing torques, pattern generator modulation, body joints),
    \item Learning rate: $1 \times 10^{-5}$,
    \item Message-Passing: 32 node channels, 4 message-passing layers,
    \item Expert policy architecture:  256-256-256 fully-connected layers with LayerNorm + Tanh (first layer) and ELU activations, outputting mean (Linear) and std (Linear + Softplus) for 12-dimensional actions.
\end{itemize}

Both tasks use identical graph topology derived from the FlyWire connectome but differ in input/output dimensions and network hyperparameters due to behavioral constraints. Training metrics are logged for performance monitoring.

\section{Detailed Settings of Locomotion Environment}
\label{app:env_details}

Our experiments follow the locomotor task design introduced in the flybody simulator, which provides imitation-learning datasets for terrestrial (\emph{walking}) and aerial (\emph{flight}) behaviors.

We evaluate four tasks:
\begin{itemize}
\setlength{\itemsep}{0pt}
\setlength{\parsep}{0pt}
\setlength{\parskip}{0pt}
    \item Gait initiation: generating stable stepping patterns from rest,
    \item Straight-line walking: tracking forward centre-of-mass (CoM) trajectories,
    \item Turning: executing lateral turns at constant speed,
    \item Flight: stabilizing and steering free flight trajectories.
\end{itemize}
For walking-based tasks, the default setting of flybody receives a 741-dimensional proprioceptive/exteroceptive observation, comprising:
\begin{itemize}
\setlength{\itemsep}{0pt}
\setlength{\parsep}{0pt}
\setlength{\parskip}{0pt}
  \item accelerometer (3), gyro (3), velocimeter (3), world $z$-axis (3),
  \item actuator activations (59),
  \item appendage pose (21), force sensors (18),
  \item joint positions (85) and velocities (85),
  \item tactile contacts (6),
  \item reference displacement (195) and root quaternion (260).
\end{itemize}
We augment this with binocular visual input: left and right eye cameras ($32\times32\times3$ RGB each) downsampled and resized to two $16\times16$ grayscale figures. The final observation dimension is 1,253.

Actions remain 59-dimensional, actuating adhesion, head/abdomen motion, and all leg joints as in default settings of flybody walking task.

For the flight task, we keep the flybody sensory design with a 104-dimensional input comprising:
\begin{itemize}
\setlength{\itemsep}{0pt}
\setlength{\parsep}{0pt}
\setlength{\parskip}{0pt}
    \item accelerometer (3), gyro (3), velocimeter (3), world $z$-axis (3),
    \item joint positions (25) and velocities (25),
    \item reference displacement (18) and root quaternion (24).
\end{itemize}
The policy outputs 12 control signals: instantaneous wing torques, head/abdomen angles, and Wing-Pattern Generator (WPG) frequency modulation. The WPG provides a nominal wing-beat template, while the policy learns residual corrections.

\section{Joint kinematics and actuator activations during walking}
\label{app:kinematics}

\begin{figure}[htbp]
  \centering
  \includegraphics[width=0.95\textwidth]{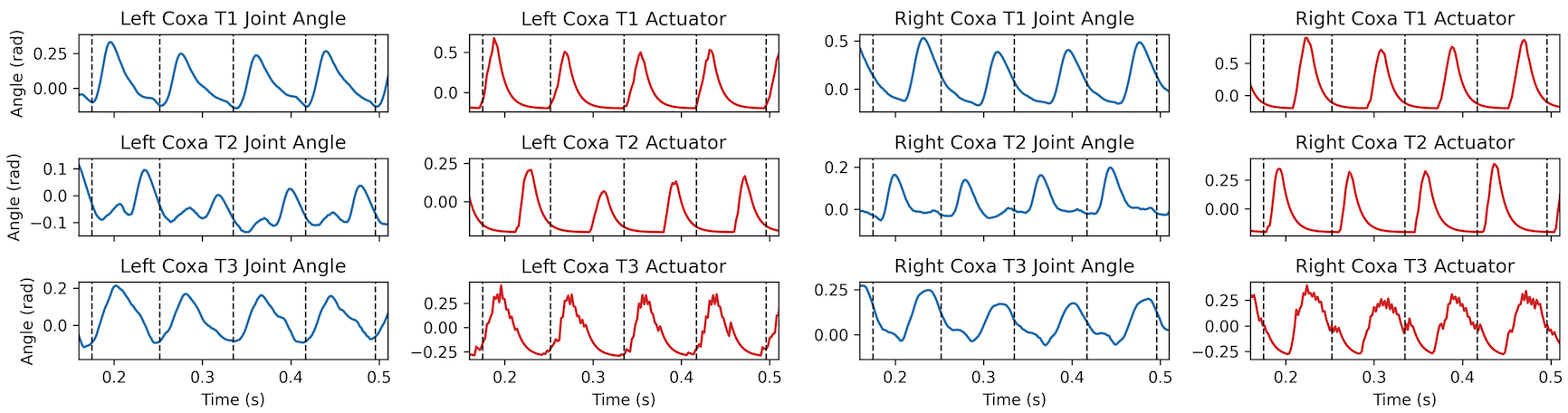}
  \caption{\textbf{Joint kinematics and actuator activations during walking.}
  We visualize the angles (blue) and actuator activations (red) of coxa joints from left (T1-T3) and right (T1-T3) legs over multiple gait cycles. Dashed vertical lines mark gait phases based on the troughs of the left T1 coxa. The model reproduces alternating tripod-like coordination, with left T1/T3 synchronized with right T2, and left T2 synchronized with right T1/T3, consistent with expected fly walking patterns.}
  \label{fig:coxa}
\end{figure}

To further investigate the motor control strategy learned by FlyGM, we analyze the joint-level kinematics and corresponding actuator activations during stable walking. As illustrated in Figure~\ref{fig:coxa}, the model exhibits rhythmic coxa joint oscillations that are tightly coupled with the underlying actuator drive. Specifically, the temporal profiles of both joint angles and activations reveal a clear alternating tripod gait: the left T1/T3 and right T2 legs form one functional group, while the right T1/T3 and left T2 legs form the other. These two groups operate in anti-phase, maintaining the static stability required for hexapedal locomotion. The synchronization between descending control signals (actuator activations) and the resulting physical movement (joint angles) demonstrates that the connectome-structured policy effectively translates brain-wide message passing into coordinated whole-body movement, consistent with the kinematic patterns observed in biological Drosophila.

\section{Detailed Method for Visualization}

\subsection{Aggregated synapse graph (Figure~\ref{fig:connectome}a)}

We visualized the aggregated synapse graph of the fly connectome using the FlyWire FAFB v783 dataset, with nodes classified by superclass labels provided by FlyWire. The node size reflects the number of neurons in each class, while the thickness and darkness of the directed edges represent the number of aggregated synaptic connections.

\subsection{Force-directed layout (Figure~\ref{fig:connectome}b)}

In the visualization of the network of connectome, we did not incorporate the three-dimensional structural priors of Drosophila neurons. Instead, we employed a force-directed layout algorithm~\citep{Kobourov2012} for graph drawing. In this method, edges are modeled as springs while nodes exert repulsive forces on each other, and the system iteratively evolves toward a low-energy equilibrium. This layout highlights the topological organization of the network, making the relationships among different super-classes and subgroups more visually apparent. For visualization, we applied a threshold of more than 25 synapses and retained only connections exceeding this cutoff.

\section{Details of Neural Representation Analysis}
\label{app:repr_analysis}

\paragraph{Reduced neural representation intensity.}
The high-dimensional features recorded during simulation have shape $\mathcal{T} \times \mathcal{N} \times \mathcal{C}$, where $\mathcal{T}$ is the number of time steps, $\mathcal{N}$ is the number of neurons, and $\mathcal{C}$ is the channel dimension. We compress these features into a single-channel signal via Principal Component Analysis (PCA) along the channel dimension. The resulting values are clipped to the 5th-95th percentile range and min-max normalized to $[0,1]$ to improve comparability and reduce outlier effects.

\paragraph{Stratified downsampling.}
Due to the extreme imbalance in superclass sizes (e.g., $>77000$ neurons in optic classes versus only 106 in motor classes), we employed stratified random downsampling to ensure visibility across all functional groups while preserving relative proportions. The downsampling thresholds were set as follows: sensory: 400, ascending: 200, optic: 1500, central: 400, visual projection: 120, visual centrifugal: 120, descending: 120, motor: 100, endocrine: 60. This approach maintained the diversity of neural responses while creating visually interpretable group representations.

\paragraph{Spectral sequencing for neuron ordering.}
To optimize the visualization of temporal dynamics, we reorder neurons within their functional classes using spectral sequencing. We first build the similarity matrix $S$ as
\begin{equation}
S_{ij} = \beta \, \text{sim}_{\text{cos}}(x_i, x_j) + (1 - \beta)\, d_{ij},
\end{equation}
where $\text{sim}_{\text{cos}}$ is the cosine similarity, $d_{ij}$ is the normalized Euclidean distance, and $\beta$ is set to 0.7. We then construct the symmetric normalized Laplacian matrix
\begin{equation}
L = I - D^{-1/2} S D^{-1/2},
\end{equation}
where $D$ is the degree matrix with $D_{ii} = \sum_j S_{ij}$. Eigendecomposition of $L$ yields eigenvalues $\lambda_0 \le \lambda_1 \le \dots \le \lambda_n$ and corresponding eigenvectors $\mathbf{v}_0, \mathbf{v}_1, \dots, \mathbf{v}_n$. The optimal ordering is determined by the Fiedler vector $\mathbf{v}_1$, the eigenvector corresponding to the second smallest eigenvalue $\lambda_1$, giving the final visualization sequence
\begin{equation}
\pi = \text{argsort}(\mathbf{v}_1).
\end{equation}
This reordering ensures that neurons with similar temporal activation patterns are positioned adjacently, revealing the emergent functional specialization across superclasses.

\subsection{Random-graph control}
\label{app:repr_rand}
\begin{figure}[htbp]
  \centering
  \includegraphics[width=0.8\textwidth]{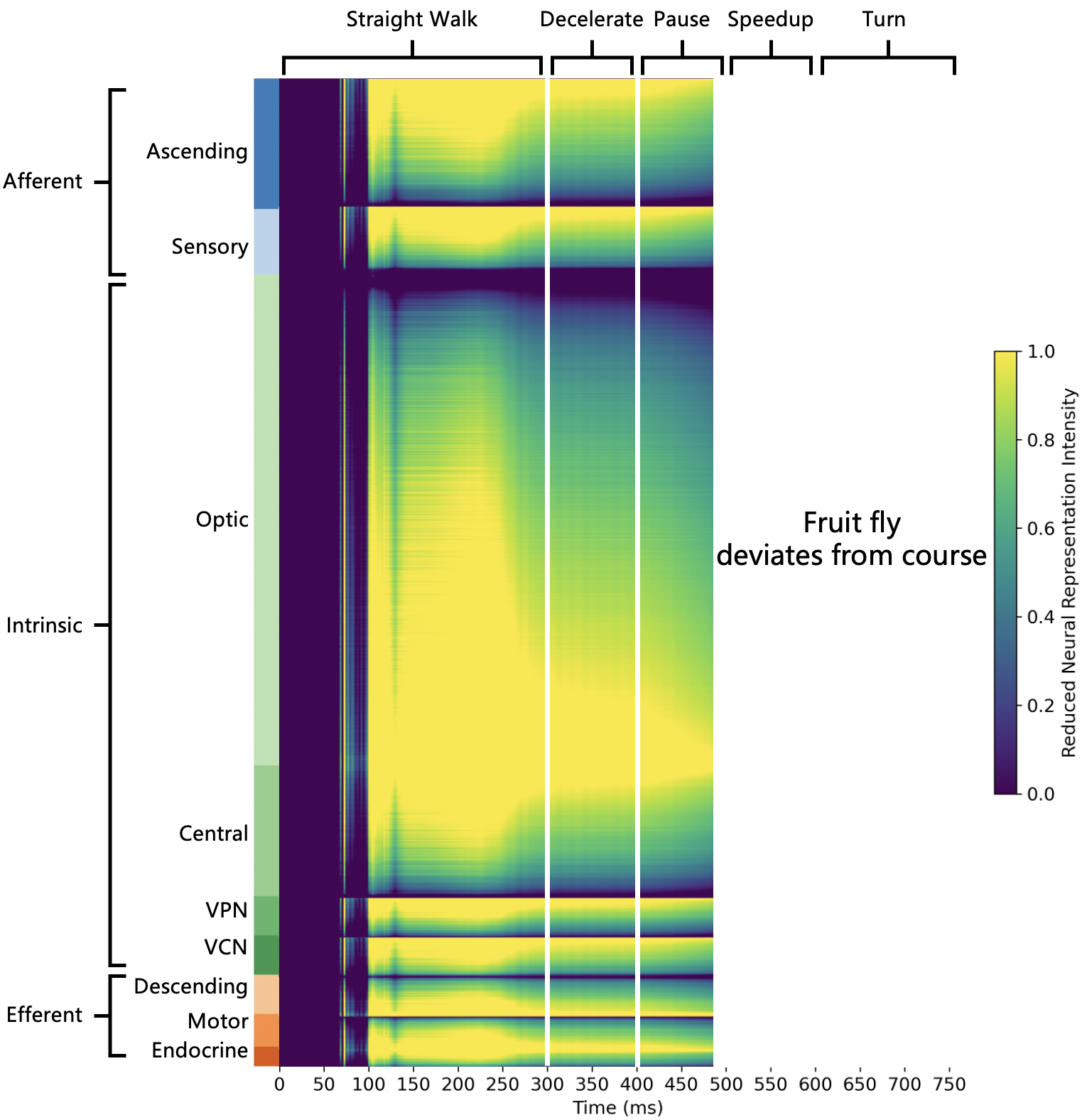}
  \caption{\textbf{Neural and behavioral dynamics of the Erd\H{o}s-R\'enyi random-graph control.} Same analysis pipeline as Figure~\ref{fig:repr_v2}, but applied to a policy whose underlying graph is an Erd\H{o}s-R\'enyi random graph with the same number of nodes and edges as the connectome. Activation patterns are largely uniform across superclasses, with no clear functional differentiation between sensory, central, and motor groups, in contrast to the structured dynamics observed in FlyGM.}
  \label{fig:repr_rand}
\end{figure}

To verify that the class-wise functional differentiation observed in Figure~\ref{fig:repr_v2} arises from connectome-specific wiring rather than from the visualization pipeline, we apply the identical analysis (PCA reduction, stratified downsampling, and spectral reordering within each superclass) to a policy trained on an Erd\H{o}s-R\'enyi random graph with matched node and edge counts. Figure~\ref{fig:repr_rand} shows the resulting dynamics. In sharp contrast to FlyGM, the random-graph model produces nearly homogeneous activation patterns across superclasses: there is no clear separation between sensory, central, and motor populations, and within-class neurons do not exhibit the coherent, behavior-locked transitions seen in the connectome-structured policy. This confirms that the emergent functional segregation in FlyGM is a consequence of the biological wiring diagram rather than an artifact of the analysis or the partition of neurons into superclasses.

\section{SNN Baseline Implementation}
\label{app:snn}

The SNN baseline is built with the SpikingJelly \texttt{activation\_based} module. We use leaky integrate-and-fire (LIF) neurons with $\tau{=}2.0$ and an arctangent surrogate gradient for backpropagation through the non-differentiable spike. To isolate the effect of spike-based dynamics, the network mirrors the MLP baseline exactly and only swaps activations: \texttt{Linear(1253$\to$32) $\to$ LIF $\to$ Linear(32$\to$512) $\to$ LIF $\to$ [Linear(512$\to$512) $\to$ LIF] $\times 3$}

The network is followed by two non-spiking linear heads producing the action mean and (Softplus-activated) standard deviation. A RunningNorm layer normalizes inputs, matching the FlyGM and MLP interface.

During training the policy unrolls a sequence $[T, B, \text{obs\_dim}]$ step by step with membrane potentials accumulating across timesteps. \texttt{functional.reset\_net} clears all LIF states at the start of each sequence. At rollout the policy is queried one step at a time ($T{=}1$) with membrane potentials persisting across steps, exposing the same single-step interface as the MLP.

Distillation, optimization, and data are identical to the FlyGM and MLP baselines. The training objective is the KL divergence between the student and expert action distributions plus a weighted MSE on $\mu$ and $\sigma$ with weight $\max(0, 10(1{-}\text{step}/100))$. We use AdamW with learning rate $1{\times}10^{-4}$ and ReduceLROnPlateau (factor $0.5$, patience $5$) on validation loss. Models were trained to convergence on a single A100 GPU on the same walking imitation dataset (\texttt{walk\_100\_in\_100\_1007}, $\text{obs\_dim}{=}1253$, $\text{act\_dim}{=}59$). The best checkpoint reached an average return of $34.64$ versus $\geq 334$ for FlyGM, and never produced a stable gait.

Table~\ref{tab:snn_perf} reports the SNN's control performance under the same speed/yaw conditions as Table~\ref{tab:topology_reduction}. Errors are one to two orders of magnitude above every connectome-based or MLP baseline. Under the higher-yaw conditions ($\psi{=}4$ and $\psi{=}7$), the policy fell over at gait initiation and the episode terminated immediately, so seeds collapsed onto the same worst-case error and the std is essentially $0$. This confirms that single-neuron biological plausibility alone is insufficient to replace connectome-level structural priors.

\begin{table}[htbp]
\centering
\small
\setlength{\tabcolsep}{6pt}
\renewcommand{\arraystretch}{1.1}
\begin{tabular}{l cccc}
\toprule
\textbf{Metric} & $v{=}2,\psi{=}0$ & $v{=}3,\psi{=}0$ & $v{=}3,\psi{=}4$ & $v{=}3,\psi{=}7$ \\
\midrule
Angle Err $\downarrow$ & $94.87{\pm}20.14$ & $108.17{\pm}17.83$ & $133.17{\pm}0.00$ & $125.73{\pm}0.00$ \\
\bottomrule
\end{tabular}
\vspace{0.5em}
\caption{SNN baseline control performance under the same conditions as Table~\ref{tab:topology_reduction}. Angle error in degrees. Under high-yaw conditions the policy terminates immediately at gait initiation, so all seeds collapse onto the same worst-case error and the std is essentially $0$.}
\label{tab:snn_perf}
\end{table}

\section{FlyGM Ablations: Edge Weights and Intrinsic Descriptors}
\label{app:flygm_ablation}

We ablate two components of FlyGM. The first is the signed synaptic edge weights, comparing \emph{Weighted} against \emph{Unweighted}, where the latter sets all edges to unit strength. The second is the per-neuron intrinsic descriptors, comparing the full per-neuron descriptors against \emph{Shared Desc}, where descriptors are tied across neurons of the same superclass, and \emph{No Desc}, where descriptors are removed. Table~\ref{tab:flygm_ablation} reports control performance after the full IL+PPO pipeline using the same conditions as Table~\ref{tab:topology_reduction}.

Two observations stand out. First, the connectome is the dominant prior. Even with descriptors heavily reduced (Shared Desc) or removed (No Desc), every variant still outperforms the MLP and all non-connectome baselines from Table~\ref{tab:topology_reduction}, confirming that the topology itself drives most of the inductive bias. Second, signed edge weights provide complementary information. Weighted variants consistently match or beat their unweighted counterparts on angle error in the high-yaw condition (e.g., $8.29$ vs.\ $11.00$ for the full model at $\psi{=}7$), where precise orientation control is hardest.

\begin{table}[htbp]
\centering
\small
\setlength{\tabcolsep}{6pt}
\renewcommand{\arraystretch}{1.1}
\begin{tabular}{ll cccc}
\toprule
\textbf{FlyGM Variant} & \textbf{Metric}  & $v{=}2,\psi{=}0$ & $v{=}3,\psi{=}0$ & $v{=}3,\psi{=}4$ & $v{=}3,\psi{=}7$ \\
\midrule
{\textbf{Weighted (full)}}
& Angle Err $\downarrow$ & $4.96{\pm}0.09$ & $\mathbf{5.57{\pm}0.06}$ & $6.36{\pm}0.33$ & $\mathbf{8.29{\pm}0.21}$ \\
\midrule
{\textbf{Unweighted}}
& Angle Err $\downarrow$ & $4.91{\pm}0.13$ & $5.90{\pm}0.10$ & $6.93{\pm}0.12$ & $11.00{\pm}0.27$ \\
\midrule
{\textbf{No Desc + Weighted}}
& Angle Err $\downarrow$ & $5.26{\pm}0.12$ & $6.02{\pm}0.11$ & $7.03{\pm}0.08$ & $9.17{\pm}0.17$ \\
\midrule
{\textbf{No Desc + Unweighted}}
& Angle Err $\downarrow$ & $5.00{\pm}0.05$ & $6.59{\pm}0.13$ & $8.52{\pm}0.13$ & $13.23{\pm}0.20$ \\
\midrule
{\textbf{Shared Desc + Weighted}}
& Angle Err $\downarrow$ & $\mathbf{4.25{\pm}0.21}$ & $5.59{\pm}0.22$ & $\mathbf{6.01{\pm}0.17}$ & $9.10{\pm}0.38$ \\
\midrule
{\textbf{Shared Desc + Unweighted}}
& Angle Err $\downarrow$ & $4.75{\pm}0.09$ & $6.61{\pm}0.08$ & $8.65{\pm}0.16$ & $13.43{\pm}0.67$ \\
\bottomrule
\end{tabular}
\vspace{1em}
\caption{Ablation of edge weights and intrinsic descriptors in FlyGM, evaluated under the same conditions as Table~\ref{tab:topology_reduction}. Angle error in degrees. Values are mean$\pm$std with per-column best in bold.}
\label{tab:flygm_ablation}
\end{table}

\section{Comparison with Standard GNN Architectures}
\label{app:gnn_baselines}

To address the concern that FlyGM's advantage over MLP and random-graph baselines might be attributable to its graph-structured computation rather than specifically to its biological wiring, we compare FlyGM against five standard GNN architectures on the walking imitation and reinforcement learning pipeline: Graph Convolutional Network (GCN)~\citep{Kipf2017a}, Edge Convolution Network (EdgeCNN)~\citep{Wang2019}, Graph Attention Network (GAT)~\citep{Velickovic2018}, GraphSAGE~\citep{Hamilton2018}, and Principal Neighbourhood Aggregation (PNA)~\citep{Corso2020}. All baselines operate on the same FlyWire node partition (afferent, intrinsic, efferent) and are trained under the identical IL+PPO pipeline as FlyGM.

For each baseline, the node embedding dimension and network depth were selected as the largest configuration that fits within the memory budget of a single NVIDIA A100 80\,GB GPU. Despite this hardware-matched allocation, no GNN baseline approaches FlyGM's performance, as reported in Table~\ref{tab:gnn_baselines}.

\begin{table}[htbp]
\centering
\small
\setlength{\tabcolsep}{8pt}
\renewcommand{\arraystretch}{1.15}
\begin{tabular}{lcc cc}
\toprule
\textbf{Model} & \textbf{Dim} & \textbf{Depth} & \textbf{Eval KL} $\downarrow$ & \textbf{Avg.\ Reward} $\uparrow$ \\
\midrule
GCN         & 16 & 4 & 7.93 & 43.13 \\
EdgeCNN     & 4 & 2 & 5.54 & 90.48 \\
GAT         & 8 & 2 & 3.43 & 132.65 \\
GraphSAGE   & 16 & 4 & 3.31 & 125.55 \\
PNA         & 4 & 2 & 2.89 & 145.33 \\
\midrule
\textbf{FlyGM (ours)} & 32 & 4 & \textbf{2.34} & \textbf{334.34} \\
\bottomrule
\end{tabular}
\vspace{0.8em}
\caption{Comparison of FlyGM against standard GNN architectures on the walking task. Eval KL measures the KL divergence to the expert policy at the end of the imitation learning stage; Avg.\ Reward is the mean episodic return under the RL reward function after the full IL+PPO pipeline. Dim and Depth denote node embedding dimension and number of message-passing layers, respectively. For each baseline, these were set to the largest configuration that fits within a single A100 80\,GB GPU.}
\label{tab:gnn_baselines}
\end{table}

Two observations stand out. First, FlyGM outperforms all GNN baselines by a substantial margin on both metrics. It achieves the lowest Eval KL ($2.34$ versus $2.89$ for the next-best PNA) and more than doubles the average reward of the strongest GNN competitor ($334.34$ versus $145.33$ for PNA). Second, the performance gap persists even though FlyGM operates at a larger node dimension ($32$) than most baselines. The ablations in Appendix~\ref{app:flygm_ablation} further show that even an unweighted FlyGM at reduced capacity exceeds every non-connectome baseline. Together, these results indicate that the advantage of FlyGM does not stem solely from graph-structured computation or parameter budget, but specifically from the biological connectivity structure of the Drosophila connectome and the signed synaptic weighting scheme it encodes.


\end{document}